# Distance Oriented Kalman Filter Particle Swarm Optimizer

[1]Chengjia Wang, [2]Keith A. Goatman, [3]James Boardman, [2]Erin Beveridge, [1]David Newby, and [1]Scott Semple

[1]BHF Centre for Cardiovascular Science, University of Edinburgh, Edinburgh, [2]Toshiba Medical Visualization Systems, Edinburgh, [3]MRC Centre for Reproductive Health, University of Edinburgh, Edinburgh.

*Abstract*—In this paper we describe improvements to the particle swarm optimizer (PSO) made by inclusion of an unscented Kalman filter to guide particle motion. We demonstrate the effectiveness of the unscented Kalman filter PSO by comparing it with the original PSO algorithm and its variants designed to improve performance. The PSOs were tested firstly on a number of common synthetic benchmarking functions, and secondly applied to a practical three-dimensional image registration problem. The proposed methods displayed better performances for 4 out of 8 benchmark functions, and reduced the target registration errors by at least 2mm when registering down-sampled benchmark brain images. Our methods also demonstrated an ability to align images featuring motion related artefacts which all other methods failed to register. These new PSO methods provide a novel, efficient mechanism to integrate prior knowledge into each iteration of the optimization process, which can enhance the accuracy and speed of convergence in the application of medical image registration.

## I. Introduction

Optimization is a key component in many practical scientific computing problems. It is used to search for the optimum value of a pre-defined fitness function of a measure within a problem space [1]. As a typical global optimization method, particle swarm optimization (PSO) has been paid significant attention during last few decades, as it is less prone to becoming trapped in local optima. Various improvements have been suggested to the original PSO algorithm to improve convergence and computation speed.

However, neither the original PSO method nor its existing modifications derived any advantage from available prior knowledge about the problem space which may act as a critical role in specific applications. The goal of many optimization problems is not just searching for an optimal value of the fitness function. One typical example of this issue is presented by problem associated with image registration, for which the distance to the real global optima, rather than the value of the measurement function, is more important. This is because small fitness differences but large distances in the problem space actually represent large differences between image transformation parameters, which may in turn falsely indicate alignment between images. If prior knowledge about the content of the image is ignored in favour of the result of the value-oriented PSO, the optimization process may tend to converge to local optima that exhibit "better" measurement values. These local optima may be at a significant distance from the global optimum, thereby causing the image registration to "fail". To deal with this special type of application, in this paper, we introduce a novel distance-oriented PSO, guided by an unscented Kalman filter (UKF) [1]. This method can encode prior knowledge about the distribution of a fitness function within the problem space, and tends to stretch the optimizer to converge at a point near the true global optimum.

Image registration algorithms are often based on the premise that the magnitude of the chosen similarity metric is related to the magnitude of the error between the current spatial transform and the optimal spatial transform between the images [1, 2]. Assuming the distribution of the similarity metric function is approximately unimodal, we propose a customized UKF-PSO framework derived from the Bayesian perspective of the PSO [3]. The UKF-PSO algorithm iteratively estimates global optima with accumulated information about probability distributions of the similarity measurements. This leads to faster convergence, with improved robustness to local optima over a large search space. Another advantage of this approach is the ease with which multiple similarity metrics can be combined, by extension to a nested UKF-PSO (N-UKF-PSO) that removes the need to apply fixed weights to the different similarity metrics by adaptively adjusting the weighting during the convergence process of the Kalman filter. The proposed methods are compared to several popular PSO methods using some popular benchmark functions, as well as a publicly available medical image registration dataset. Both the UKF-

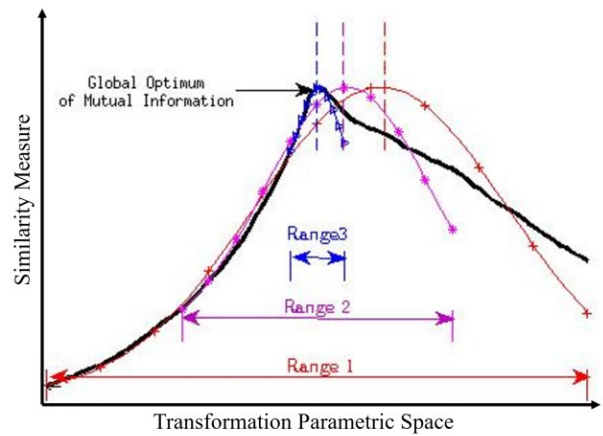

Fig. 1. Fitting a Gaussian function to the distribution of mutual information within three different searching ranges. The Gaussian function tends to give a more accurate estimation of the distribution within a smaller searching range.

PSO and N-UKF-PSO display better robustness to local optima and better accuracies in the image registration experiments.

In this paper, important previous work that attempts to solve similar image registration problems using the original or modified versions of PSO are briefly reviewed in section II. The theory of our UKF-PSO and N-UKFPSO methods are introduced in section III. Sections IV and V describe the details of UKF-PSO and N-UKF-PSO. Experiments performed on both benchmark functions and a publicly available image registration dataset are shown in sections VI and VII, and discussed in section VIII.

## II. RELATED WORKS

Both local and global optimization methods have been applied to solve image registration problems. Local optimization suffers from becoming trapped in local optima. The use of multi-resolution image pyramids can partially mitigate this, however, the global optimum may not be represented in the down-sampled problem spaces, in which case the optimizer will still converge to a local optima [4]. Among the global optimization methods, evolutionary computation plays an important role. For example, inspired by social and cooperative behavior, Kennedy and Eberhart [5] proposed the first PSO algorithm in the mid-1990s [6]. Since then a number of modified versions of PSO have been developed and applied to different image registration applications [4, 7]. Research efforts have concentrated on improving the convergence speed and robustness of the PSO when the problem spaces are very large and exhibit multiple local optima. These extensions of PSO methods use either alternative neighbourhood structures [8] or novel particle evolution strategies [6, 7, 9]. A widely used PSO using alternative particle evolution formulae is quantum behaved PSO (QPSO) [9]. The formulae were further redesigned in the revised QPSO (RQPSO), the diversity controlled RQPSO (DRQPSO) [10] and the chaotic search

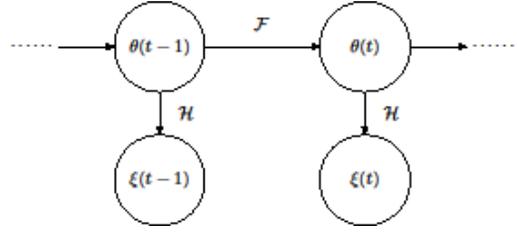

Fig. 3. Hidden Markov model: $\theta$ and $\xi$ are the hidden and observed states.

QPSO [11]. Another popular approach is to hybridize PSO with other optimization methods, for example Genetic algorithm [12] or Simplex [13]. Comparisons and reviews of the major PSO variants can be found in [3].

Wachowiak's method provides a registration-specific prior knowledge approach [4], but requires precise initialization. Other methods that exploit prior knowledge include the Bare Bones PSO [14], Kalman Filter PSO [15], and Andras' Gaussian PSO, based on a Bayesian interpretation [3]. These methods either provide a probabilistic perspective of the particle status, or a mechanism to integrate prior knowledge.

## III. THEORY DERIVATION

For a fitness function $f(x)$, an optimization process search in a problem space $\Omega$ for a $x$ that gives optimal value of $f(x)$. For the problem targeted by this paper, image registration, $f(x)$ is a predefined similarity measure between images, and $\Omega$ is all the possible image transformations limited by degrees of freedom. The purpose of optimization is then formulated by:

$$x^o = arg\max_{x \in \Omega} f(x), \quad (1)$$

where $x^o$ is the optimal solution of $x$, and the purpose of registration is to find $x^o$ which gives the optimal image transformation parameters, or leads to the highest similarity of the images. However, due to the presence of local optima, $x^o$ is often difficult to find. In this case, the returned $x$ should be as close as possible to $x^o$.

The PSO simulates the social and cooperative behavior of a "swarm" of potential solutions, called particles [6]. Each potential solution corresponds to one position in problem space. Each particle explores the problem space at an individual random speed that is partially affected by combined knowledge about the up-to-date global and local optima. Searching for global optima in a $D$-dimension problem space with $K$ particles at the $t$th iteration of PSO, a solution represented by the position of the $i$th particle is a $D$-element vector, $\mathbf{x}_i(t) = \{x_{i1}(t), x_{i2}(t), \cdots, x_{iD}(t)\}, i \in \{1, 2, \cdots, K\}$. In the original PSO method, a widely used formula for updating the speeds of the particles, $v_i(t+1)$, is given by [3, 5, 6]:

$$v_i(t+1) = \omega v_i(t) + c_p r_p \left(x_i^p - x_i(t)\right) + c_g r_g (x^g - x_i(t)) \quad (2)$$

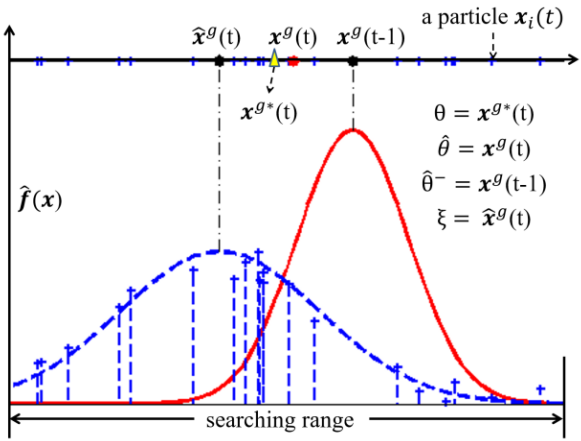

Fig. 2. Information available at the $t$th iteration of PSO: the hidden state $\theta$ represents an optimal estimation $x^{g*}$ of the true global optimum; the observed state $\xi$ is defined as the average position of all particles $\hat{x}^g$ weighted by the measured fitness function $\hat{f}$ of each particle. An estimation of the hidden state $x^g$ is produced by fitting $\hat{f}$ to a Gaussian function in each iteration of the optimization process. For the $t$th iteration, $x^g(t)$ can be obtained by combining $x^g(t-1)$ and $\hat{x}^g(t)$. When solving the optimization problem using a linear Kalman filter, $x^g(t-1)$ is treated as the output of $time - update$ stage, $\hat{\theta}$-, and $x^g(t)$ is the output of the $measurement - update$ stage, $\hat{\theta}$.

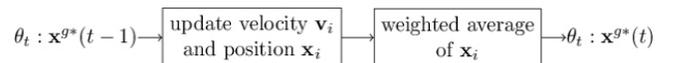

Fig. 4. The non-linear state transition model $\mathcal{F}$ used to evolve the optimal estimation $x^{g*}$ of the true global optimum.

where $\omega$ is the inertia weight, $x_i^p$ is the local best solution found by the $i$th particle, and $x^g$ is the best up-to-date global optimum. $c_p$ and $c_g$ are acceleration constants that weight the attraction of local and global optima to each particle, and $r_p$ and $r_g$ are random generated numbers drawn from the uniform distribution over the range of (0,1) [6]. The updated particle positions are then given by [3, 5, 6]:

$$x_i(t+1) = x_i(t) + v_i(t+1). \qquad (3)$$

Equation (2) consists of three components: the previous velocity $v_i(t)$, the cognition component $c_p r_p \left(x_i^p - x_i(t)\right)$, and the social component $c_g r_g (x^g - x_i(t))$. The combination of these components is a compound velocity that moves the particles towards the local and global optima, while preventing any significant deviations from the particles' previous directions [6]. This mechanism makes a stepwise improvement in the algorithm convergence until all of the particles have moved into a small constrained area, or the global best position remains unchanged for a certain number of iterations. Other than the coefficients which appear in the PSO formula, the most common modifiable parameters are the swarm size (i.e. the number of particles), the searching range, and the maximum number of iterations.

If $f(x)$ is complicated and presents multiple local optima which is common for image registration applications, PSO still suffers from premature convergence. Integration of prior knowledge of the problem space into the particle evolution formulae can improve the robustness of PSO. Andras [3] proposed a Gaussian PSO model based on a Bayesian interpretation. In this model, the evaluated fitness value, $\hat{f}(x)$ [3] is given by:

$$\hat{f}(x) = f(x) + \epsilon, \qquad (4)$$

where $\epsilon$ is a noise distribution (typically zero-mean Gaussian) added to the noise-free fitness value [3]. Following Bayesian theory, likelihood is given in the form of a probability density function (PDF), $\mathcal{P}(x)$, defined over the search range. Given all $\hat{f}(x_i(0))$, the PDF may be calculated using:

$$\mathcal{P}\left(x \middle| \hat{f}(x_i(0))\right) = \frac{\mathcal{P}(\hat{f}(x_i(0))|x) \cdot \mathcal{P}(x)}{\mathcal{P}\left(\hat{f}(x_i(0))\right)}, i = 1, \cdots, K \quad (5)$$

where the posterior $\mathcal{P}\left(x \middle| \hat{f}(x_i(0))\right)$ calculated in an iteration is used as the new $\mathcal{P}(x)$ in the following iteration. The evolution of the particles can then be formulated by [3]:

$$x_i(t+1) = x_i(t) + \gamma \cdot \frac{\partial}{\partial x} \ln \mathcal{P}_{t+1}(x) \Big|_{x = x_i(t)}, \qquad (6)$$

where $\mathcal{P}_t(x)$ is the $\mathcal{P}(x)$ calculated in the $t$-th iteration. The calculation of $\mathcal{P}_t(x)$ can be performed based on the assumption that the evaluated fitness values of the particles are either co-dependent or independent, leading to two implementations of this Bayesian Gaussian PSO. The fitness function is assumed to be proportional to the probability of a point in the search range being the optimal solution. Thus in a registration problem, the similarity measure can be considered as a non-normalized probability. The probability distribution over the whole search range is interpolated using multiple Gaussian bases.

Andras [3] provides a framework to integrate prior knowledge into image registration in the form of $\mathcal{P}(x)$ [3]. In this paper, we use a simplified definition of $\mathcal{P}(x)$, based on prior knowledge specific to image registration. As a result, there is no need to calculate the probability distribution under different assumptions of dependences between particles, as $\mathcal{P}(x)$ can be directly fitted using the evaluation of all particles.

Target registration error (TRE) is often the ground truth metric of image registration problems. The TRE is zero for two perfectly aligned images. We generalize this, such that $x^o$ is the optimal transformation represented as a point in the problem search space, that results in a TRE closest to zero. Over the whole search range, the similarity measure $f(x_i)$ of a transformation represented by any particle is the distance measure $\|x_i - x^o\|$. Any other similarity measure can be considered as a monotonic mapping of this distance, $K(\|x_i - x^o\|)$. We simply assume a form of Gaussian function for $K$,

$$f(x_i) = exp\left(-\frac{\beta}{2}\|x_i - x^o\|^2\right). \qquad (7)$$

This assumption of prior knowledge indicates that $\mathcal{P}(x)$ follows a Gaussian-like distribution with unknown expectation, $x^o$. The advantage of using this Gaussian form is that $x^o$ equals to the expectation $\int x \mathcal{P}(x) \, dx$ over the whole problem space. In each iteration of the PSO, $x^o$ is estimated by the optimum value $x^g$ within the area searched by particles. Here, rather than directly selecting the optimum value from among all particles, the estimated global optimum $x^g(t)$ is calculated by the average of all $x_i(t)$ weighted by the normalized $\hat{f}(x_i(t))$ defined in equation (4).

According to (4), and the theory of the Bayesian interpretation of the PSO [3], $\mathcal{P}(x)$ is thus modeled as:

$$\mathcal{P}(x_i) = \sigma \cdot \left(exp\left(-\frac{\beta}{2}\|x_i - x^g\|^2\right) + \epsilon\right), \qquad (8)$$

where $\sigma$ is a normalization constant and $\epsilon$ is a zero-mean Gaussian noise with unknown standard deviation. Ignoring the noise $\epsilon$, a reasonable estimation of $\mathcal{P}(x)$ is:

$$\hat{\mathcal{P}}_t(x_i) = \sigma \cdot \left(exp\left(-\|x_i - \hat{x}^g(t)\|^2 \Big/ 2\hat{\delta}^2\right)\right), \qquad (9)$$

where the $\hat{\mathcal{P}}_t(x_i)$ is the estimation of $\mathcal{P}(x)$ at $x_i$ in the $t$th iteration. $\hat{\mathcal{P}}_t(x)$ can be obtained by fitting a Gaussian function using all $\hat{f}(x_i(t))$. $\hat{\delta}^2$ is the variance of this Gaussian function. The global optimum can be estimated by solving,

$$\frac{\partial}{\partial x}\hat{\mathcal{P}}_t(x) = 0. \qquad (10)$$

Although the assumed Gaussian form of $\hat{f}(x)$ and $\hat{\mathcal{P}}(x_i)$ cannot accurately capture the shape of the similarity measure for a large search range, it gives a reasonable estimation of the global optima, and will improve as the search range contracts, as shown in Fig. 1. If the searching algorithm converges ideally, the Gaussian function becomes a Dirac delta function.

Equation (10) can be solved by fitting the shape of $\ln \hat{\mathcal{P}}_t(x)$ using a quadratic least squares method, though this will introduce much greater computational complexity. The purpose of fitting the Gaussian function is to obtain an estimated global optimum $\hat{x}^g(t)$, and $\hat{\delta}^2$ is not used in further optimization processes. We use the weighted mean of all particles obtained in each iteration to estimate the initial global optimum, i.e.

$$\hat{x}^g(t) = \left(\sum_{i=1}^K x_i(t)\hat{f}(x_i(t))\right) \bigg/ \left(\sum_{i=1}^K \hat{f}(x_i(t))\right). \quad (11)$$

The estimation of the global optimum should move towards the true global optimum of the similarity measure as the search range contracts during the optimization process. One important assumption of (10) is that $\hat{f}(x) \geq 0$, which is easy to achieve by normalization. Specific to image registration problems, if the images are aligned by minimizing a difference measure, denoted as $\hat{f}^d(x)$, we can convert it to a similarity measure by,

$$\hat{f}(x) = exp\left(-\varepsilon\left(\hat{f}^d(x)\right)\right) \quad (12)$$

where $\varepsilon(\cdot)$ is a function of $\hat{f}^d(x)$ in the search range.

In summary, during each iteration of the PSO, a noisy estimation of the global optimum $\hat{x}^g(t)$ can be obtained using (11). $\hat{x}^g(t)$ can then be improved during the evolutionary process of the PSO by combining information from all of the particles and all of the previous iterations.

## IV. THE LDS-KFPSO METHOD

$\hat{x}^g$ calculated using (11) can replace $x^g$ in the PSO formulae as it moves closer to the optimum of $\hat{f}(x)$. However, with integrated prior knowledge, the estimation of the PDF of $x^o$ in the search range can be improved by accumulating the information obtained in previous iterations. This can be achieved through the dynamic Bayesian network (DBN) presented in Monson and Seppi's Kalman filter PSO [15], which is used to characterize the time-sensitive relationship between observable and hidden states. For image registration problems using swarm optimization, the global and local optima obtained in each iteration can be encoded as the observed state $\xi$. Based on the theory in [15], the desired hidden state $\theta$ represents the ideal location and speed of a particle that leads to a better fitness of $x^{g*}$. With the prior knowledge discussed above we can define $x^{g*}$ as the average of $x_i$, weighted by the noise-free fitness function, $f(x_i)$, or more directly define it as $x^{g*} = x^o$. An estimation $\hat{\theta}$ of the hidden state is given for each iteration.

However, because the prior knowledge of registration problems is integrated and $\hat{x}^g$ is calculated using equation (11), a much simpler DBN can be adopted here, using the raw information demonstrated in Fig. 2. After $t-1$ iterations, the hidden state is the ideal position $x^{g*}(t)$ that is closer to $x^o$, or equals $x^o$. The observation $\xi$ can be directly defined as $\hat{x}^g(t)$. Each iteration has a current estimation of the hidden state $x^{g*}(t)$ based on this observation. To obtain this estimation, the relationship between $\theta$ and $\xi$ is depicted as an instance of the hidden Markov model (HMM), as shown in Fig. 3 [15]. The hidden state $\theta$ evolves over time, based on a state transition model $\mathcal{F}$, and influences the observable state through a known observation model $\mathcal{H}$. The transition model, $\mathcal{F}$, reflects how an estimated global optimum moves closer to locations of better fitness, and the observation model can then be described as a model of the influence of $x^{g*}(t)$ upon $\hat{x}^g(t)$. When defining $x^{g*}$ as the average of $x_i$ weighted by $f(x_i)$, as shown in Fig. 4, $\mathcal{F}$ can be specified such that the evolution of $x^{g*}$ depends on the movements of every particle. This assumes either a highly non-linear state transition process, or we may use $x^o$ as the hidden state that assumes an identical state transition. In both cases, the observation model is an identical mapping.

This influence of $x^{g*}$ on $\hat{x}^g$ is inherently noisy, and the noise is used as a subjective uncertainty model of the accuracy of an observation [15]. Based on the prior knowledge being integrated, the current state is modeled by a Gaussian distribution with mean $x^g$ and a variance that models how strong the likelihood is that $x^g$ reflects $x^{g*}$. The goal of the registration process is then to reduce the uncertainty of this likelihood over $x^g$ to its lowest level, and thus give the most accurate prediction. Since this prediction is produced by combining the information from all particles and all previous iterations, it is applicable to different PSO methods with different velocity and position updating mechanisms.

For the HMM described above, the Kalman filter [16] and its extensions [2, 17] can be regarded as solutions. When $\mathcal{F}$ and $\mathcal{H}$ are linear, and the HMM is therefore known as a linear dynamic system (LDS), the Kalman filter provides an efficient way to recursively estimate the state of this process while minimizing the mean square error [18]. The Kalman filter models the HMM as a *predictor-corrector* circle, where both the state-transition and observation are noisy processes with additive Gaussian noise. In our registration problem assuming a LDS in the prediction or *time-update* stage, a prediction of $x^{g*}(t)$ is given by,

$$\hat{\theta}^-(t) = \mathbf{F}\hat{\theta}(t-1), \quad (13)$$

$$\mathbf{\Sigma}^-(t) = \mathbf{F}\mathbf{\Sigma}(t-1)\mathbf{F}^T + \mathbf{\Sigma}_\theta, \quad (14)$$

where $\mathcal{F}$ is the matrix representation of the state transition function, $\theta^-(t)$ and $\mathbf{\Sigma}^-(t)$ are the mean and variance of predicted $x^g(t)$ respectively, and $\mathbf{\Sigma}_\theta$ is the covariance of the state-transition noise. Assuming $\theta(t) = x^{g*}(t) = x^o$, $\mathbf{F}$ is an identity matrix. Then in the *correction*, or *measurement-update* stage, the estimation of state is refined using the observation,

$$\mathbf{K}(t) = \frac{(\mathbf{F}\mathbf{\Sigma}(t-1)\mathbf{F}^T + \mathbf{\Sigma}_\theta)\mathbf{H}^T}{\mathbf{H}(\mathbf{F}\mathbf{\Sigma}(t-1)\mathbf{F}^T + \mathbf{\Sigma}_\theta)\mathbf{H}^T + \mathbf{\Sigma}_\xi}, \quad (15)$$

$$\hat{\theta} = \hat{\theta}^-(t) + \mathbf{K}(t)\left(\xi(t) - \mathbf{H}\hat{\theta}^-(t)\right), \quad (16)$$

$$\mathbf{\Sigma}(t) = (\mathbf{I} - \mathbf{K}(t)\mathbf{H})\mathbf{\Sigma}^-(t), \quad (17)$$

where the $\mathbf{K}(t)$ is the Kalman gain in the $t$th iteration that is used to balance the influence of prediction and observation, $\mathbf{H}$ is the observation matrix, which is identity, and $\hat{\theta}(t)$ and $\mathbf{\Sigma}(t)$ are the mean and variance of the estimation respectively. The estimate of global optimum is based on the following probability distribution [18],

$$\mathcal{P}(\theta(t)|\xi(t)) \sim N\left(\hat{\theta}(t), \mathbf{\Sigma}(t)\right). \quad (18)$$

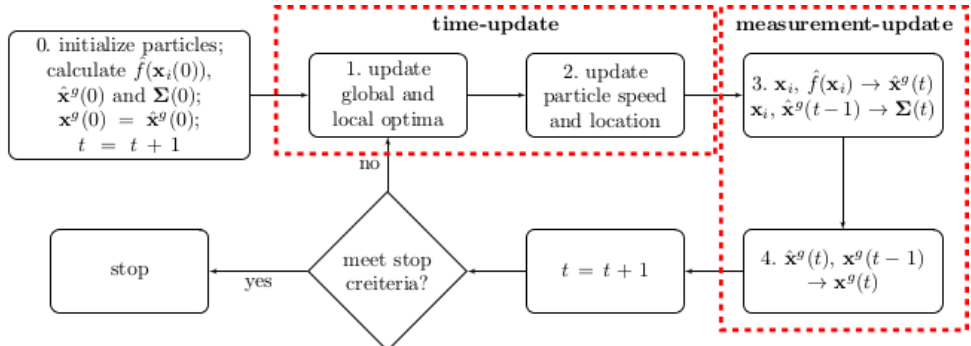

Fig. 5. Brief workflow of the unscented Kalman filter particle swarm optimizer (UKFPSO).

This PSO model guided by Kalman filter (KF) under LDS assumption is named as LDS-KFPSO.

## V. THE SPO-UKFPSO METHOD

When using the non-linear state transition model shown in Fig. 4, the HMM is not a LDS. In this case the non-linear extensions of the Kalman filter should be applied to deal with the non-linear state transition process $x^{g*} = \mathcal{F}(x^{g*}(t-1))$. The extended Kalman filter (EKF) is the standard method for dealing with non-linear processes. However, it requires the calculation of a Jacobian matrix for $\mathcal{F}(x)$ [2], which is difficult for this complicated state transition function. Hence we propose the novel use of an unscented Kalman filter (UKF) [2]. Rather than estimate an arbitrary transition function as the EKF does, the UKF approximates a Gaussian probability distribution using standard vector and matrix operations based on a set of weighted sigma points, $\chi(t-1), j = 1, \cdots, 2D+1$ [19]. For the $t$th iteration in a $D$-dimensional problem space, the sample mean and covariance of the set of sigma points are $\hat{\theta}(t-1)$ and $\mathbf{\Sigma}(t-1)$ [19]. Specifically, the sigma points and their associated weights are selected by,

$$\chi_j(t-1) = \begin{cases} \hat{\theta}(t-1), j=0; \\ \hat{\theta}(t-1) + \sqrt{(D+\kappa)\mathbf{\Sigma}(t-1)}, \\ \quad j=1,\cdots,D; \\ \hat{\theta}(t-1) - \sqrt{(D+\kappa)\mathbf{\Sigma}(t-1)}, \\ \quad j=D+1,\cdots,2D+1; \end{cases} \quad (19)$$

$$\mathbf{W}_j = \begin{cases} \kappa/(D+\kappa), j=0, \\ 1/(2(D+\kappa)), j=1,\cdots,2D+1. \end{cases} \quad (20)$$

where $\mathbf{W}_j$ is the weight associated with the $j$th sigma point. Details of how to select the weighting parameter, $\kappa$, can be found in [2] and [19]. In this work, we follow Uhlmann's [19] recommendation that $\kappa + D = 3$. In the Kalman update stage each sigma point is instantiated through the state transition function by [19],

$$\chi_j(t|t-1) = \mathcal{F}\left(\chi_j(t-1)\right), \quad (21)$$

and then the mean of state prediction is calculated by [19]:

$$\theta^-(t) = \sum_{j=0}^{2D} \mathbf{W}_j \chi_j(t|t-1), \quad (22)$$

and the variance is given by [19],

$$\mathbf{\Sigma}^-(t) = \sum_{j=0}^{2D} \mathbf{W}_j = \left(\chi_j(t|t-1) - \theta^-(t)\right) \cdot \left(\chi_j(t|t-1) - \theta^-(t)\right)^T. \quad (23)$$

As the observation model is an identity function, we can still use the linear measurement update formulae of the original

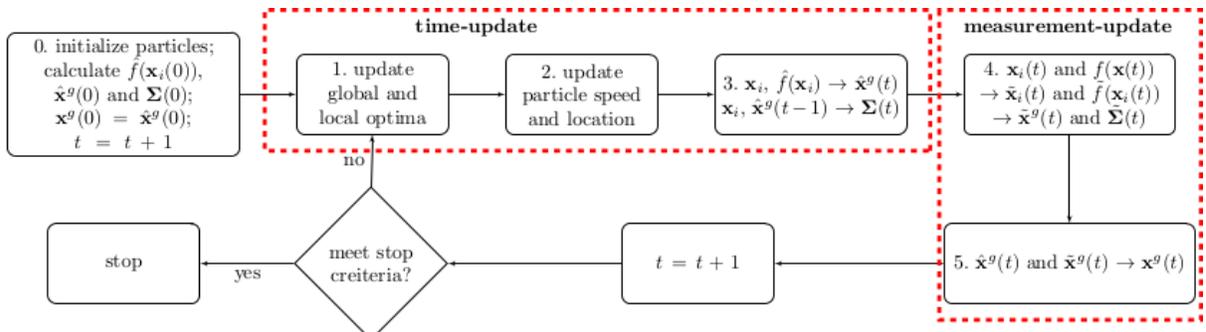

Fig. 6. Workflow of the unscented Kalman filter particle swarm optimizer with "shift particles observation" (SPO-UKFPSO).

Kalman filter (given by equations (16-18)) in the correction stage to obtain $\theta(t)$ and $\Sigma(t)$.

Under this non-LDS assumption, since the uncertainty associated with the estimated global optimum is related to the distribution of particles, we can simply use either a sample, or all of the particles together with the estimated global optimum as the sigma points of UKF. This allows the number of sigma points to be greater than $2D + 1$, and makes integrating the UKF into the PSO more convenient. In addition to the traditional stopping criteria, $\Sigma(t)$ may be used as additional evidence of the convergence situation of the PSO. To sum up, the procedure of the PSO was combined with the predict-correct circle of the Kalman filter. For both LDS and non-LDS cases, our new UKF-PSO algorithm can be represented as shown in Fig. 5.

The estimated global optimum, $\hat{x}^g$, will be affected by the relative location of the global optimum in the search range. The estimation is more accurate when the true global optimum is closer to the center of the search range. A slightly different observation can therefore be used to improve the estimated global optimum: in each iteration, after $\hat{x}^g$ is calculated, all the particles are resampled to be $\tilde{x}_i$, so that the searching range is centered on $\hat{x}^g$. Then a new average $\tilde{x}^g$ can be calculated as the observation, weighted by the new evaluations $\tilde{f}(\tilde{x}_i)$. We name this model the "shift particles observation" UKFPSO (SPO-UKFPSO). In this case, the HMM will be different from the one used in the above UKFPSO method, with different definitions of $\theta$, $\hat{\theta}$, $\Sigma$, $\mathcal{F}$, and $\xi$. The workflow of the SPO-UKFPSO method is shown in Fig. 6. To apply the UKF guided PSO model to real image registration tasks, the choice of similarity measure also has a profound influence on the results. The chosen similarity measure has to follow the prior knowledge modeled by equation (8-10), which allows the problem to be solved as shown in Fig.1. For example, for a multi-modality registration problem, the sum of squared difference (SSD) of intensity is a poor choice. Therefore, we opt for the widely used mutual information (MI) instead. To register a reference image $\mu$ and a floating image $\nu$, MI is calculated using their joint entropy $H(\mu, \nu)$, and marginal entropies $H(\mu)$ and $H(\nu)$,

$$MI(\mu, \nu) = H(\mu) + H(\nu) - H(\mu, \nu), \quad (24)$$

where MI makes registration a maximization problem.

## VI. THE NESTED UKF-PSO

Image registration can be performed using different types of similarity measures, as well as different features. In order to combine different features and measures we must assign a suitable weighting to each one, and normalize them to comparable scales. A benefit of the proposed model using prior knowledge, is that fitness values of any similarity measure are automatically normalized so as to be samples of a probability distribution, which maps all the measures to a uniform scale.

As shown in Fig. 7, in the case where we have two similarity measures, $f_1(x)$ and $f_2(x)$, the estimation of the global optimum output by a UKF using one measure can be intuitively considered as $\theta^-$ of the second UKF associated with the other measure. The two UKFs share the same population of particles during the optimization process, which means that each particle obtains two fitness values in each iteration. The framework can be extended using multiple nested UKFs to allow any number of features or similarity measures to guide the optimization.

## VII. PARTICLE STATE EVOLUTION OF UKF GUIDED PSO

To sum up, the outputs of the KF or UKF in the three implementations of PSO above include the estimated hidden state $\hat{\theta}$, and a variance $\Sigma$, that reflects the estimation error. As discussed in sections III and IV, the accuracy of the estimation of the global optimum given by the weighted average (equation (11)) is dependent on the size of the search region, and the positioning of the true global optimum. Furthermore, the KF and its extensions generally behave like low-pass filters, which means high frequency information may be filtered out as well as the noise. In this case, a more reliable rapid model can be formulated by:

$$\begin{aligned} v_i(t+1) &= \omega v_i(t) + c_p r_p \left(x_i^p - x_i(t)\right) \\ &+ c_g r_g (x^g - x_i(t)) + c_\theta r_\theta \left(\hat{\theta} - x_i(t)\right), \end{aligned} \quad (25)$$

where $c_\theta$ is the acceleration constant weighting the attraction of the estimated hidden state output by the KF or UKF, and $r_\theta$ is a randomly generated number drawn from the uniform distribution over the range $(0, 1)$. The component $c_\theta r_\theta (\hat{\theta} - x_i(t))$ introduced in equation (25) controls the influence of the estimated hidden state over the orientation of particles. The acceleration constants $c_p$, $c_g$ and $c_\theta$ need to be adjusted to balance the influence of the personal optima $x_i^p$, the measured global optimum $x^g$, and the filtered optimum $\hat{\theta}$. Many methods initialise $c_p$ and $c_g$ as 2.0. In this work, $c_g$ and $c_\theta$ are initialized by letting $c_g = c_\theta = 1$, and during the particle evolution process they are adjusted by

$$c_g = min(\|x_g(t) - x_g(t-1)\|, 1.2), \quad (26)$$

and

$$c_g = 2 - c_\theta, \quad (27)$$

where $x^\theta(t)$ is the measured global optimum $x^g$ obtained in the $t$th iteration. Particle positions are then updated using equation (3).

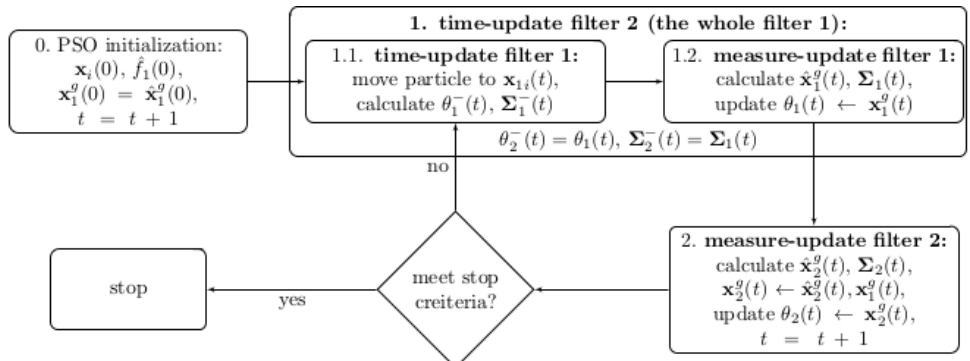

Fig. 7. Workflow of nested unscented Kalman filter particle swarm optimizer (nested-UKFPSO).

## VIII. EXPERIMENTS

The proposed PSO methods were evaluated on both general optimization and image registration problems. A few representative PSO methods previously used for registration are also chosen for comparison purposes.

### A. Benchmark Functions

The proposed PSO models were compared using some common benchmark functions widely used in the PSO literature [20], shown in table I. Since the optimization methods proposed in this paper are customized for image registration applications with the assumed prior knowledge described in section IV, we chose different types of benchmark functions, both single-objective and multi-objective, to comprehensively compare the power of the different PSO methods. As the nested UKFPSO method is specifically designed for image registration applications requiring multiple types of features or different types of similarity measures, it is not included in this benchmark function comparison.

For image registration problems, it is more important to find a position that is closer to the real global optima in the search space than to search for a better value of the fitness function. The performances of the compared algorithms are therefore measured by the norm of the differences between their returned vectors and the ground truths of the benchmark functions. Since for most of the chosen benchmark functions the ground truth optima locate in the center of the search space, a weak optimization algorithm that tends to converge to the center of search space may obtain better results than others. To deal with this bias, while keeping the ground truth within the search space, we generated random shifts of the searching bounds, limited to be within 40% of the problem space.

Besides the random shift of the search ranges, the algorithms were tested using a random problem dimension chosen between 2 to 30, and repeated for each algorithm 100 times for each benchmark function. The mean and standard deviation (STD) of each algorithm were calculated. The stop condition of the algorithms was either reaching 300 iterations or reduction of the variability of the particle positions around the global optima to be less than $10^{-6}$. All of the algorithms were implemented in MATLAB (Mathworks, USA) with vectorized simulation of particle positions. Other than the particle position update mechanism, and some method specific parameters, all of the different implementations shared core code to ensure that the comparison was performed under similar circumstances.

Accuracy, convergence speeds and the run times of each method were measured. Speeds were evaluated using the average number of iterations and function evaluations of each run, as well as the raw convergence time. For a general overview of the performances, the mean accuracy of each method over all benchmark functions was also calculated.

TABLE I
BENCHMARK FUNCTIONS

| Function Name | Ackley | Griewank | Modulus Sum | Rastrigin |
|---|---|---|---|---|
| $f(x) =$ | $-20 \cdot exp\left(-0.2 \cdot \sqrt{\frac{1}{D}\sum_{d=1}^{D} x_d^2}\right) - exp\left(\frac{1}{D}\sum_{d=1}^{D} \cos(2\pi x_d)\right) + 20 + e$ | $\frac{1}{4000}\sum_{d=1}^{D} x_d^2 - \prod_{d=1}^{D} \cos\left(\frac{x_d}{\sqrt{d}}\right) + 1$ | $60 + \sum_{d=1}^{D} |x_d|$ | $100 + \sum_{d=1}^{D}(x_d^2 - 10 \cdot \cos(2\pi x_d))$ |
| Bounds | $[-30, 30]^D$ | $[-600, 600]^D$ | $[-5.12, 5.12]^D$ | $[-5.12, 5.12]^D$ |
| Ground Truth | $(0, 0)^D$ | $(0, 0)^D$ | $(0, 0)^D$ | $(0, 0)^D$ |
| 1D-plots | 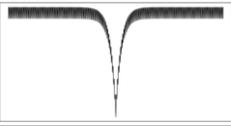 | 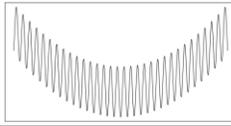 | 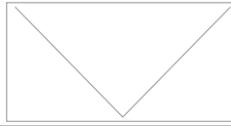 | 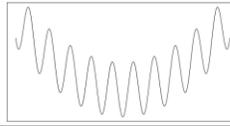 |
| Function Name | Salomon | Schwefel | Rosenbrock | Step |
| $f(x) =$ | $1 - \cos\left(2\pi\sqrt{\sum_{d=1}^{D} x_d^2}\right) + 0.1\sqrt{\sum_{d=1}^{D} x_d^2}$ | $5000 + \sum_{d=1}^{D} -x_d \sin(\sqrt{|x_d|})$ | $\sum_{d=1}^{D-1}((x_d - 1)^2 + (x_{d+1} - x_d^2)^2 \cdot 100)$ | $60 + \sum_{d=1}^{D}\lfloor x_d \rfloor$ |
| Bounds | $[-100, 100]^D$ | $[-500, 500]^D$ | $[-30, 30]^D$ | $[-5.12, 5.12]^D$ |
| Ground Truth | $(0, 0)^D$ | $(420.968746, 420.968746)^D$ | $(1, 1)^D$ | $(-5.12, 5.12)^D$ |
| 1D-plots | 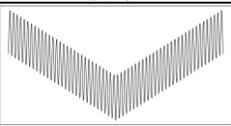 | 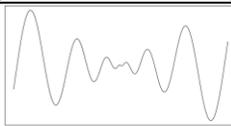 | 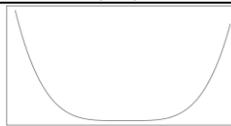 | 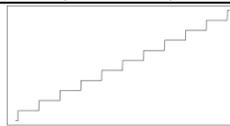 |

The variable $x$ is a $D$-dimension vector with the form $(x_1, x_2, \cdots, x_D)$.

## B. Registering Benchmark Datasets

In order to evaluate the performances of the proposed PSO methods in real registration applications, we conducted a rigid registration experiment based on data from the multi-modality brain image datasets from the Retrospective Image Registration Evaluation (RIRE) Project [21]. The comparison includes the original PSO, the DRQPSO, the Bare Bones PSO, the Kalman filter PSO, LDS-KFPSO, SPO-UKFPSO and the nested UKFPSO methods. All methods use MI as the similarity measure, except for the nested UKFPSO, which used MI for measure $f_1(x)$ and the gradient features proposed by Pluim et al. [22] were used as $f_2(x)$.

We performed CT-MR_T2 and PET-MR_PD registration. The voxel size is $0.65×0.65×4mm^3$ for CT data, $1.25×1.25×4mm^3$ for MR_T2 and MR_PD data, and $2.59×2.59×8mm^3$ for PET data.

As the purpose of this experiment is to compare the performance of different PSO methods in real image registration applications, rather than to obtain the absolute highest registration accuracy, we integrated the PSO methods into a very simple registration framework. For the sake of simplicity and efficiency, each slice of both the reference and floating volumes was down-sampled to 20% of the original in-plane resolution of the reference image along each dimension. The slice thickness of the floating volume was also interpolated to the slice thickness of the reference volume so that the optimization method only dealt with translation and rotation parameters. To allow further speed-up of the registration, we selected a cubic region of interest (ROI) in each volume by applying Otsu's histogram-based threshold selection method [23] to the normalized data. The RIRE project measures the accuracy of registration using TRE, calculated from multiple volumes of interest (VOIs). TRE is used as the measure of registration accuracy. The transformation parameters calculated from the resampled data are rescaled for transformation of the original volume. For each patient, 10 attempts at registration were completed, and in each run all methods use the same set of initialized particles that were generated by a MATLAB quasi-random number simulator.

## C. Registering Neonatal Datasets

To further compare the performance of our methods with the original PSO, we also conducted an experiment using neonatal data collected from a clinical trial performed at the Clinical Research Imaging Centre (CRIC), University of Edinburgh (UoE). This dataset has previously been used to evaluate the performance of the registration framework based on a rearranged histogram specification (RHS) and K-means binning [24]. We used images acquired at 38-44 weeks' postmenstrual age in natural sleep using a 3T Verio system (Siemens Healthcare Gmbh, Erlangen, Germany). Because of the neonatal age of the population being imaged, there is likely to be significant motion between acquisitions, which makes this dataset a good test of registration algorithms. Isotropic anatomical data were acquired with a range of contrasts, selected to facilitate the development of volumetric brain segmentation algorithms for the main study.

Data from 10 patients were aligned using a rigid-body transform, calculated within a $51×51×41mm^3$ user-positioned ROI on volumes with an isotropic voxel size of 1.56 mm. Transformation matrices were obtained from data down-sampled to half original resolution. Performance was evaluated by TREs, calculated from 1908 pairs of corresponding landmarks (18 on each volume), manually placed by a clinical expert. The accuracy of the LDS-KFPSO and the nested-UKFPSO are compared with the results from our earlier work based on the original PSO [24].

## IX. RESULTS

### A. Benchmark Functions

Table II shows the average minimization error of the different algorithms for each benchmark function (the STD of each run is shown within parenthesis). Table III summarizes the overall performances of the different algorithms.

As shown in table II, the original PSO gave the best result for the Step function. The Bare Bones PSO performed better for the Griewank, Modulus Sum and Salomon functions. The proposed LDS-KFPSO method converged to positions that are closer to the true global optima for the Ackley Schewefel and Rosenbrock functions. For the majority of the benchmark functions, the proposed LDS-KFPSO and SPO-UKFPSO returned the best performances, or performances comparable to Bare Bones PSO. The Step function is a special case among all the benchmark functions, as it is increases monotonically, and the global optimum is located around the upper bound of the search range. In registration applications, this may happen when the true global optimum is not included in the search space. As expected, in this case, the LDS-UKFPSO and SPO-UKFPSO methods gave worse results.

TABLE II
PERFORMANCES OF THE PSO METHODS APPLIED TO THE CHOSEN BENCHMARK FUNCTIONS

| Function | Original PSO | QPSO | RQPSO | DRQPSO | Chaotic PSO | Bare bones PSO | Kalman filter PSO | LDS-KFPSO | SPO-UKFPSO |
|---|---|---|---|---|---|---|---|---|---|
| Ackley | 8.891(5.4) | 6.859(4.1) | 7.319(5.0) | 5.991(4.4) | 10.23(5.8) | 9.070(5.5) | 6.140(3.4) | **0.665(0.3)** | 1.431(1.0) |
| Griewank | 3.042(1.2) | 6.635(4.3) | 2.081(1.0) | 1.696(0.7) | 15.49(11.3) | **1.496(0.9)** | 4.652(2.3) | 1.641(0.9) | 1.616(0.8) |
| Modulus Sum | 0.013(0.03) | 0.235(0.2) | 0.005(0.01) | 0.002(0.01) | 0.786(0.6) | **8e-7(1e-6)** | 0.062(0.1) | 0.076(0.02) | 0.067(0.02) |
| Rastrigin | 0.586(0.6) | 0.614(0.5) | 0.255(0.2) | 0.270(0.2) | 1.046(0.7) | 0.334(0.3) | 0.513(0.3) | 0.239(0.03) | **0.156(0.01)** |
| Salomon | 0.536(0.5) | 4.897(4.6) | 0.620(0.8) | 0.932(1.2) | 15.01(11.9) | **0.324(0.2)** | 2.160(2.6) | 1.542(1.0) | 1.560(0.9) |
| Schwefel | 331.4(218) | 403.7(200) | 337.0(198) | 280.1(112) | 244.7(174) | 316.2(218) | 367.3(214) | **231.3(90)** | 233.4(90) |
| Rosenbrock | 0.795(0.8) | 1.754(1.7) | 0.804(0.4) | 0.857(0.5) | 5.987(2.4) | 1.455(1.3) | 1.190(0.4) | **0.567(0.13)** | 0.590(0.2) |
| Step | **0.077(0.05)** | 0.079(0.05) | 0.087(0.07) | 0.319(0.4) | 1.184(1.2) | 0.078(0.05) | 0.788(1.0) | 2.294(1.08) | 1.910(1.1) |

The performances are measured with mean and standard deviation (STD) of the distances between the returned function values and the ground truths of all benchmark functions. The mean values are shown within the parenthesis. Best results for the benchmark functions are shown in bold font.

TABLE III
EVALUATION OF THE PSO METHODS APPLIED TO THE CHOSEN BENCHMARK FUNCTIONS

| Function | Original PSO | QPSO | RQPSO | DRQPSO | Chaotic PSO | Bare bones PSO | Kalman filter PSO | LDS-KFPSO | SPO-UKFPSO |
|---|---|---|---|---|---|---|---|---|---|
| Error Per Function | 1.9917 | 3.0105 | 1.5960 | 1.4380 | 7.1051 | 1.8225 | 2.2148 | **1.0033** | 1.0466 |
| Overall Error STD | 3.2096 | 3.0312 | 2.6198 | 2.0844 | 6.5266 | 3.2557 | 2.3081 | 0.8276 | **0.7569** |
| Number of Iterations Per Run | 148.85 | 88.29 | 65.373 | 138.01 | 46.31 | 144.96 | 93.02 | **39.65** | 39.92 |
| Function Evaluation Per Run | 10804 | 6074 | 4478 | 10374 | 2994 | 10807 | 6575 | **2572** | 5054 |
| Seconds Per Run | 0.6523 | 0.3924 | 0.2919 | 0.6813 | **0.1915** | 0.6654 | 0.4213 | 0.2905 | 0.3990 |

The best result in term of each statistical criterion is shown in bold font.

TABLE IV
EVALUATION OF THE PSO METHODS APPLIED TO RIRE DATA

| Modality | Function | Original PSO | DRQPSO | Bare bones PSO | Kalman filter PSO | LDS-KFPSO | SPO-UKFPSO | SPO-UKFPSO |
|---|---|---|---|---|---|---|---|---|
| CT-MR_T2 | Mean | 6.2158 | 4.5297 | 10.3678 | 5.5092 | 3.5898 | 1.7407 | **1.1829** |
|  | Median | 6.2047 | 4.4752 | 12.0473 | 5.6158 | 3.5980 | 1.8617 | **1.1718** |
|  | STD | 2.2740 | 0.8503 | 3.6121 | 1.0642 | 1.0607 | 0.6932 | **0.3326** |
|  | Run Time | 112.33s | 97.69s | 92.40s | **73.58s** | 83.68s | 135.89s | 138.67s |
| PET-MR_PD | Mean | 3.5883 | 3.9001 | 3.6822 | 6.2004 | 3.5112 | 3.1409 | **2.9810** |
|  | Median | 3.1755 | 3.5118 | 3.7254 | 6.1185 | 3.1472 | 3.1971 | **3.0962** |
|  | STD | 1.0313 | 6.2657 | **0.3322** | 1.6303 | 1.5786 | 0.8860 | 1.0489 |
|  | Run Time | 105.94s | 78,12s | 106.67s | **59.83s** | 78.03s | 96.91s | 108.31s |

The performances are measured with mean and standard deviation (STD) of the distances (measured in mm) between the returned function values and the ground truths of all benchmark functions. The mean values are shown within the parenthesis. Best results for the benchmark functions are shown in bold font.

Based on the results shown in table III, due to the simplicity of its position update model, the implementation of chaotic QPSO has the fastest convergence time, but worst accuracy. In comparison, the LDS-KFPSO and SPO-UKFPSO take slightly longer to complete each iteration, but both required fewer iterations than other methods. In particular, the LDS-KFPSO used the least number of function evaluations, and had the shortest run time to achieve the best optimization results. The SPO-UKFPSO provided greater accuracy compared to LDS-KFPSO and converged quicker than most of the other methods.

*B. RIRE Data*

The TREs for the CT-MR_T2 and PET-MR_PD registrations are shown in table IV. All three proposed PSO methods returned better results than the other methods in terms of mean and median TRE. Due to the combined features and similarity measures it utilizes, the nested-UKFPSO gave better results amongst the three proposed PSO models. For the Bare Bones PSO and Kalman filter PSO, since these methods feature a more deterministic position update mechanism, they display better convergence speed than the original PSO and DRQPSO. However, the original PSO and DRQPSO were highly sensitive to particle initialization, and gave the greatest variability in each run of the experiment.

*C. Neonatal Data*

Fig. 8 displays the results of successfully registering the T2-w dark fluid and T1-w MRPAGE neonatal images using the UKFPSO methods. Registration of this particular dataset was only achieved using the UKFPSO method, previous methods had failed to register the shown example. The quantitative evaluation of these registration results are shown in table V. The LDS-KFPSO and nested UKFPSO therefore not only gave smaller TREs than the original PSO, but also successfully aligned one particular problematic dataset that our previous method failed to register [24].

## X. CONCLUSION

In this paper, we have described three new UKF-guided registration-oriented optimization implementations. The new PSO-based methods were evaluated using benchmark functions and by registering two medical image cohorts. Compared to the selected PSO algorithms, the UKF-guided PSO methods achieved more accurate registration results, and displayed better robustness to the presence of local optima. The convergence speed is comparable to the QPSO when minimizing benchmark functions, and is comparable to the original PSO algorithm when registering medical images.

This new type of UKF-based PSO algorithm provides an efficient mechanism to encode prior knowledge of the search

TABLE V
STATISTICS OF TARGET REGISTRATION ERRORS (TRE)

|  | Original PSO | LDS-KFPSO | nested-UKFPSO |
|---|---|---|---|
| Mean | 3.25 | 2.80 | **2.72** |
| Median | 1.88 | 1.93 | **1.82** |
| STD | 3.41 | 1.71 | **1.56** |
| Number of Failures | 1 | 0 | **0** |
| Average Run Time | 92.48s | 100.19s | 144.83s |

Errors measured in millimeter (mm).

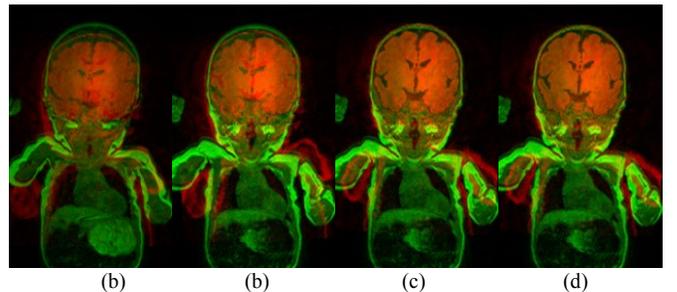

(b)    (b)    (c)    (d)

Fig. 8. Registration results obtained using original particle swarm optimizer (PSO), the linear dynamic system Kalman filter PSO (LDS-KFPSO) and the nested unscented Kalman filter PSO (UKFPSO): (a) before registration; (b) registered using original PSO; (c) registered using LDS-KFPSO; (d) registered using nested UKFPSO. The registration is performed to align the T2 weighted dark fluid and T1 MRPAGE images visualized in overlapped red and green color channels.

space into the optimization process, without requiring manually assigned weights for each feature included in that prior knowledge. Unlike other PSO methods, the proposed methods update the probabilistic distribution of the whole search space, rather than storing the distribution for each particle. This process iteratively moves the particles close to the global optimum, especially in the early stage of PSO, thus leading to quicker convergence. Furthermore, the mechanism that updates the knowledge of the search space can also be applied to other population based optimization methods, for example, other swarm intelligence methods. Thus, it has great potential for application in a variety of medical image registration problems.